\documentclass{article}

\PassOptionsToPackage{numbers, compress}{natbib}
 \usepackage[preprint]{neurips_2026}

 \makeatletter
\renewcommand{\@notice}{}
\makeatother

\usepackage[utf8]{inputenc} 
\usepackage[T1]{fontenc}    
\usepackage{hyperref}       
\usepackage{url}            
\usepackage{booktabs}       
\usepackage{amsfonts}       
\usepackage{nicefrac}       
\usepackage{microtype}      
\usepackage{xcolor}         

\usepackage{microtype}
\usepackage{graphicx}
\usepackage{subcaption}
\usepackage{colortbl,booktabs} 

\usepackage{hyperref}

\usepackage{amsmath}
\usepackage{amssymb}
\usepackage{mathtools}
\usepackage{amsthm}

\usepackage{multirow}
\usepackage{bm}
\newtheorem{problem}{Problem}
\usepackage{subcaption}
\usepackage{wrapfig}

\usepackage{calc}
\newcommand{\model}[2]{%
\makebox[1.5cm][l]{#1}#2%
}

\usepackage[capitalize,noabbrev]{cleveref}

\hypersetup{
	colorlinks=true,
	citecolor=[rgb]{0,0.08,0.45},
}

\theoremstyle{plain}
\newtheorem{theorem}{Theorem}

\theoremstyle{definition}

\theoremstyle{remark}

\title{Arena as Offline Reward: Efficient Fine-Grained Preference Optimization for Diffusion Models}

\author{Zhikai Li\textsuperscript{1}\ , Yue Zhao\textsuperscript{2}\thanks{Co-second authors.}\; , Edward Zhongwei Zhang\textsuperscript{2$*$}\ , Xuewen Liu\textsuperscript{1}\ , Jing Zhang\textsuperscript{1}\ , \\ 
\textbf{Qingyi Gu\textsuperscript{1}\thanks{Corresponding author.}\; , Zhen Dong\textsuperscript{3}} \\
\textsuperscript{1}Institute of Automation, Chinese Academy of Sciences\\
\textsuperscript{2}University of California, Berkeley \quad
\textsuperscript{3}University of California, Santa Barbara\\
}

\begin{document}

\maketitle

\begin{abstract}
Reinforcement learning from human feedback (RLHF) effectively promotes preference alignment of text-to-image (T2I) diffusion models.
To improve computational efficiency, direct preference optimization (DPO), which avoids explicit reward modeling, has been widely studied. However, its reliance on binary feedback limits it to coarse-grained modeling on chosen–rejected pairs, resulting in suboptimal optimization.
In this paper, we propose ArenaPO, which leverages Arena scores as offline rewards to provide refined feedback, thus achieving efficient and fine-grained optimization without a reward model.
This enables ArenaPO to benefit from both the rich rewards of traditional RLHF and the efficiency of DPO.
Specifically, we first construct a model Arena in which each model’s capability is represented as a Gaussian distribution, and infer these capabilities by traversing the annotated pairwise preferences.
Each output image is treated as a sample from the corresponding capability distribution.
Then, for a image pair, conditioned on the two capability distributions and the observed pairwise preference, the absolute quality gap is estimated using latent-variable inference based on truncated normal distribution, which serves as fine-grained feedback during training.
It does not require a reward model and can be computed offline, thus introducing no additional training overhead.
We conduct ArenaPO training on Pick-a-Pic v2 and HPD v3 datasets, showing that ArenaPO consistently outperforms existing baselines.
\end{abstract}

\section{Introduction}
Text-to-image (T2I) diffusion models, such as DALLE~\cite{betker2023dalle}, Stable Diffusion~\cite{rombach2022high,podell2023sdxl}, FLUX~\cite{flux2024}, and PixArt~\cite{chen2023pixart}, have demonstrated strong capabilities in generating high-quality and highly realistic images~\cite{yang2023diffusion}. Although large-scale pretraining endows these models with powerful generative abilities, post-training optimization is typically required to better align them with human preferences for real-world applications~\cite{wallace2024diffusion,xue2025dancegrpo}. Consequently, in recent years, preference alignment approaches that incorporate human feedback into the diffusion training have been widely investigated.

\begin{wrapfigure}{r}{0.51\textwidth}
  \vspace{-0.3cm}
  \begin{center}
    \centerline{\includegraphics[width=0.50\columnwidth]{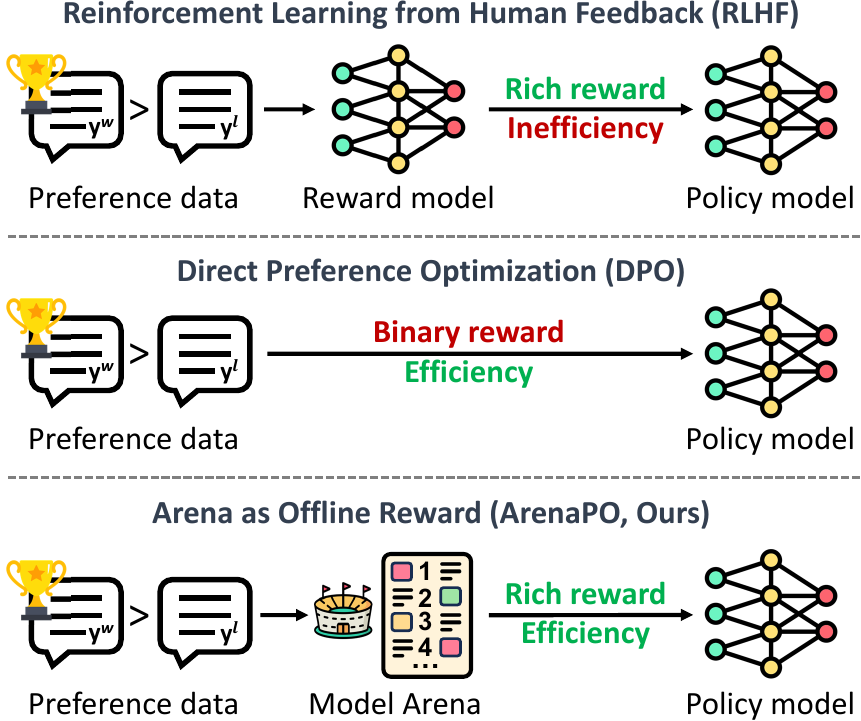}}
    \caption{
      Comparison of RLHF, DPO, and ArenaPO. The proposed ArenaPO extracts and leverages rich offline rewards from model Arena, without requiring a reward model, enabling both fine-grained and efficient preference alignment.}
    \label{fig:1}
  \end{center}
  \vspace{-0.8cm}
\end{wrapfigure}

One notable alignment paradigm is reinforcement learning from human feedback (RLHF)~\cite{christiano2017deep,stiennon2020learning}. However, its efficiency has been widely criticized, largely because the reward model requires substantial resources for training and its deployment necessitates online inference.
Therefore, most works on diffusion model alignment focus on direct preference optimization (DPO)~\cite{wallace2024diffusion,li2024aligning,hong2024margin,zhu2025dspo}, a more efficient paradigm that directly leverages static offline datasets and does not require explicit reward modeling.
Despite its high efficiency, DPO is limited to binary rewards on chosen–rejected image pairs, unlike the rich reward signals used in traditional RLHF, as illustrated in Figure \ref{fig:1}. In practice, however, the quality differences between image pairs vary substantially~\cite{li2025k}, thus this coarse-grained reward modeling and optimization can potentially lead to suboptimal results.

\begin{wrapfigure}{r}{0.51\textwidth}
  \vspace{-0.3cm}
  \begin{center}
    \centerline{\includegraphics[width=0.50\columnwidth]{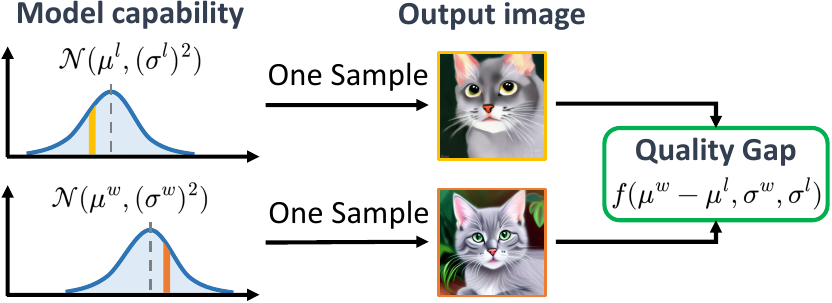}}
    \caption{
     Motivation of the proposed ArenaPO. Each output image can be viewed as one sample drawn from the model’s capability distribution $\mathcal{N}(\mu, \sigma^2)$. Thus, for two output images, their quality gap can be modeled as a function with respect to their source models' $\mu^w-\mu^l$, $\sigma^w$, and $\sigma^l$.
     }
    \label{fig:2}
  \end{center}
    \vspace{-0.8cm}
\end{wrapfigure}

In this work, we aim to explore how fine-grained preference can be incorporated into DPO in an offline manner without relying on an explicit reward model. Following Thurstone’s Case V theory~\cite{thurstone1927law}, we model each model’s capability as a Gaussian distribution characterized by a mean and uncertainty, rather than a fixed point, and regard a specific output image as a sample drawn from this distribution. The labels in the dataset represent discretized binary relative preferences between the output images.
Under this formulation, the difference between the model capability means serves as a \emph{prior of overall quality}, while the binary preference acts as an \emph{observation induced by capability uncertainty}. 
Consequently, the absolute quality gap between two output images can be estimated by a function with respect to the source models’ capability means and uncertainties, as illustrated in Figure \ref{fig:2}.

\begin{figure}[t]
  \begin{center}
    \centerline{\includegraphics[width=0.99\textwidth]{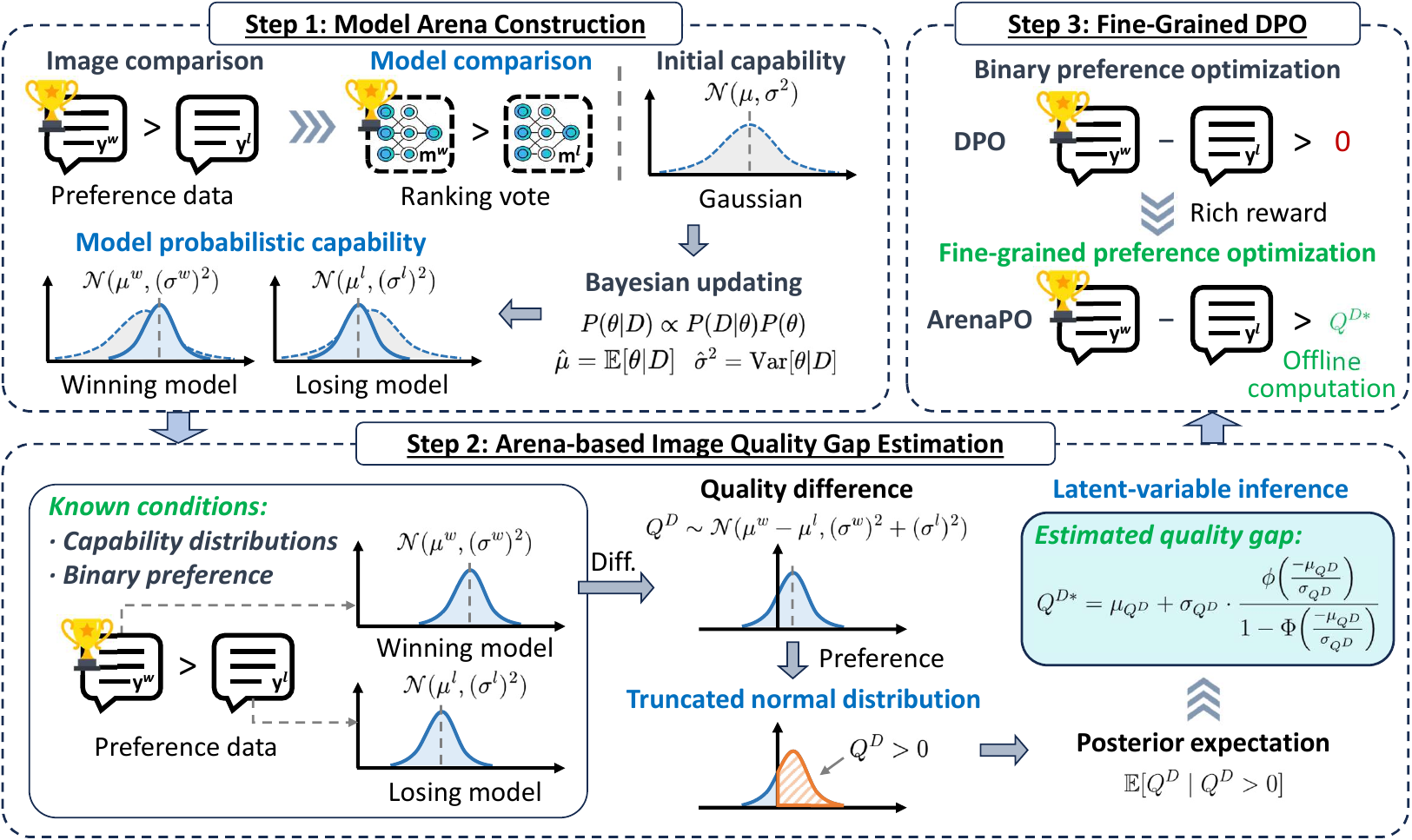}}
    \caption{
      Overview of the proposed ArenaPO. First, we build a model Arena that converts image-level pairwise comparisons in the dataset into model-level comparisons, and updates each model’s capability distribution using Bayesian inference. Second, for each data instance, the two images are regarded as samples from their corresponding capability distributions. Conditioned on the two distributions and the observed binary preference, we estimate the quality gap using latent-variable inference via truncated normal distribution. Finally, the estimated quality gap serves as an offline, fine-grained reward to enable more effective preference optimization.}
    \label{fig:overview}
  \end{center}
    \vspace{-0.4cm}
\end{figure}

With the above insight, we propose ArenaPO, which leverages Arena information to provide fine-grained preference in an offline manner.
ArenaPO can benefit from both the rich reward signals of traditional RLHF and the efficiency of DPO.
Specifically, as in K-Sort Arena~\cite{li2025k}, we construct a model Arena and represent each model’s capability as a Gaussian distribution. By traversing the entire dataset and performing pairwise comparisons, we iteratively update the capability distribution of each model.
Then, for each image pair, given the corresponding capability distributions of the two models and the annotated binary preference, we introduce a latent-variable inference procedure based on truncated normal distribution to efficiently estimate the absolute quality difference between the two images. This estimated difference serves as a fine-grained preference during training.
Notably, the entire pipeline does not rely on a reward model and can be computed fully offline, thereby incurring no additional training overhead. An overview of the proposed method is illustrated in Figure \ref{fig:overview}.

Our main contributions are summarized as follows:
\begin{itemize}
    \item We propose ArenaPO, which extracts fine-grained preference from model Arena and uses it as efficient feedback for DPO, thereby effectively improving alignment performance for T2I diffusion models. ArenaPO does not rely on a reward model and can be computed offline, thus delivering both accuracy and efficiency.
    \item We introduce a latent-variable inference method based on truncated normal distribution, which enables efficient estimation of the absolute quality difference between an image pair given the models’ capability distributions and the annotated binary preference.
    \item ArenaPO training is conducted on Pick-a-Pic v2 and HPD v3 datasets, and the results across multiple T2I benchmarks demonstrate ArenaPO's superiority. For instance, with ArenaPO training, the win-rate reaches 79.0\% compared to original Stable Diffusion 1.5.
\end{itemize}

\section{Related Work}

\textbf{T2I Diffusion Models }
T2I diffusion models have become the dominant paradigm due to their strong generation quality~\cite{croitoru2023diffusion,rombach2022high}. Recent works focus on improving model capacity and conditioning quality, including more powerful backbones~\cite{peebles2023scalable,si2024freeu}, stronger text encoders~\cite{chen2023pixart}, and more detailed captions~\cite{chatterjee2024getting}. Advances in diffusion formulations and training objectives, such as improved parameterizations~\cite{song2019generative,karras2022elucidating} and flow-based variants~\cite{lipman2022flow,liu2022flow}, further enhance performance.
Despite this, these pre-trained models typically require post-training to perform alignment optimization for real-world applications~\cite{wallace2024diffusion}.

\textbf{Human Preference Alignment }
Preference optimization was initially developed for large language models (LLMs) to align the outputs of pre-trained models with human preferences~\cite{christiano2017deep,stiennon2020learning}. In the traditional RLHF paradigm, human preferences are first used to train a reward model, and the target model is then optimized via reinforcement learning by maximizing the reward scores~\cite{schulman2017proximal}. However, this pipeline has been widely criticized for its inefficiency. Fortunately, DPO~\cite{rafailov2023direct} provides a more efficient alternative by eliminating the need for a reward model and relying solely on offline preference dataset, and has thus been widely studied~\cite{azar2024general,yuan2024self}.

\textbf{Alignment of Diffusion Models }
Preference alignment for diffusion models has recently attracted increasing attention~\cite{xue2025dancegrpo,karthik2025scalable}. Most existing methods are built upon the DPO paradigm. For instance, Diffusion-DPO~\cite{wallace2024diffusion} reformulates likelihood optimization as error modeling, resulting in a tractable training objective. DSPO~\cite{zhu2025dspo} leverages score matching to align the post-training objective with the pre-training objective. DMPO~\cite{li2025divergence} performs preference alignment by optimizing the reverse KL divergence. MaPO~\cite{hong2024margin} directly optimizes the likelihood margin between image pairs, yielding a reference-free approach.
Despite its efficiency advantages, DPO is limited by discrete binary preferences and cannot benefit from the rich reward signals in traditional RLHF, thus leading to suboptimal optimization. In this work, we introduce fine-grained preferences extracted from Arena into DPO, enabling both efficient and more effective alignment.

\section{Preliminaries}

\textbf{Diffusion Models }
Diffusion models generate samples by learning to invert a progressive noise corruption process~\citep{ho2020denoising,song2020score}. Let $p_{\text{data}}$ denote the data distribution. Starting from a clean sample $\mathbf{x}_0 \sim p_{\text{data}}$, the forward diffusion process gradually injects Gaussian noise over $T$ steps according to a predefined noise schedule $\beta_1, \dots, \beta_T$. At an arbitrary timestep $t$, the resulting sample $\mathbf{x}_t$ follows:
\begin{equation}
\begin{gathered}
q\left(\mathbf{x}_t \mid \mathbf{x}_0\right)=\mathcal{N}\left(\sqrt{\bar{\alpha}_t} \mathbf{x}_0,\left(1-\bar{\alpha}_t\right) I\right), \\
\bar{\alpha}_t=\prod_{s=1}^t \alpha_s, \quad \alpha_t=1-\beta_t.
\end{gathered}
\end{equation}
To synthesize data, the reverse (denoising) dynamics are modeled using a neural network $\epsilon_\theta$, which is trained to predict the injected noise $\epsilon_t$ from a corrupted sample $\mathbf{x}_t$ at timestep $t$. Learning is carried out by minimizing a weighted denoising score matching objective~\citep{ho2020denoising}:
\begin{equation}
\mathcal{L}_{\mathrm{DM}}(\theta)
=\mathbb{E}_{\mathbf{x}_0,\epsilon,t,\mathbf{x}_t}\Big[\omega(\lambda_t)\,
\|\epsilon-\epsilon_\theta(\mathbf{x}_t,t)\|_2^2\Big],
\end{equation}
where $\lambda_t = \log(\alpha_t^2/\sigma_t^2)$ denotes the signal-to-noise ratio, and $\omega(\lambda_t)$ is a predefined weighting function.

\textbf{DPO for Diffusion Models }
DPO~\citep{rafailov2023direct} provides a simple framework for optimizing language models directly from preference data without relying on an explicit reward model. Given a prompt $\mathbf{c}$ and a preference-labeled pair $(\mathbf{x}_0^w, \mathbf{x}_0^l)$, DPO models preferences with the Bradley-Terry (BT) model~\cite{bradley1952rank} as follow:
\begin{equation}
    p_{\mathrm{BT}}\left(\mathbf{x}_0^{w} \succ \mathbf{x}_0^{l} \mid \mathbf{c}\right)=\sigma\left(r\left(\mathbf{c}, \mathbf{x}_0^{w}\right)-r\left(\mathbf{c}, \mathbf{x}_0^{l}\right)\right),
\end{equation}
where $r(\mathbf{c}, \mathbf{x}_0)$ is the latent reward, and $\sigma$(·) is the sigmoid function. 
DPO optimizes the target policy against a fixed reference policy, with the reward implicitly given by their log-likelihood ratio as follows:
\begin{equation}
\begin{aligned}
\mathcal{L}_{\mathrm{DPO}}(\theta)
=-\mathbb{E}_{\mathbf{c},\mathbf{x}_0^w, \mathbf{x}_0^l} \Big[
\log \sigma\Big(
\beta \log \frac{\pi_\theta(\mathbf{x}_0^w | c)}{\pi_{\mathrm{ref}}(\mathbf{x}_0^w | c)} 
-\beta \log \frac{\pi_\theta(\mathbf{x}_0^l | c)}{\pi_{\mathrm{ref}}(\mathbf{x}_0^l | c)}\Big) \Big],
\end{aligned}
\end{equation}
where $\pi_\theta$ is the target policy to be optimized, and $\pi_{\mathrm{ref}}$ is a fixed reference policy (typically the pre-trained model). Here, log~$\sigma$(·) is the log-sigmoid function, and the hyperparameter $\beta$ controls the optimization strength. 

Diffusion-DPO~\citep{wallace2024diffusion} extends the DPO framework to diffusion models by interpreting the diffusion denoising objective $\mathcal{L}_{\text{DM}}$ as an implicit reward signal. The key idea is to quantify, at each timestep $t$, how much the fine-tuned model $\epsilon_\theta$ improves over a reference model $\epsilon_{\text{ref}}$ in terms of denoising quality. Accordingly, a timestep-dependent reward for a sample $x$ is defined as:
\begin{equation}
    \delta_\theta(\mathbf{c}, \mathbf{x}_t, t)=-\Big(\left\|\epsilon-\epsilon_\theta(\mathbf{x}_t, \mathbf{c}, t)\right\|_2^2-\left\|\epsilon-\epsilon_{\mathrm{ref}}(\mathbf{x}_t, \mathbf{c}, t)\right\|_2^2\Big) .
\end{equation}
Building on this per-timestep reward formulation, the resulting preference learning objective is obtained by aggregating contributions across the entire diffusion trajectory:
\begin{equation}
\begin{aligned}
    \mathcal{L}_{\text {Diff-DPO}}(\theta)=-\mathbb{E}_{\mathbf{c},\mathbf{x}_0^w, \mathbf{x}_0^l,t}\Big[  \log \sigma\Big(  
     \beta T \omega(\lambda_t)
     (\delta_\theta(\mathbf{c}, \mathbf{x}^{w}, t) -\delta_\theta(\mathbf{c}, \mathbf{x}^{l}, t))\Big)\Big],
\end{aligned}
\end{equation}
where $T$ denotes the total number of diffusion steps, and $\omega(\lambda_t)$ is a weighting function that controls the relative importance of different timesteps.

\section{Methodology}

In this chapter, we first construct an Arena to obtain each model's capability distribution in section \ref{sec:arena}; then, we perform problem formulation for image quality gap estimation and propose a solution based on latent-variable inference, as detailed in section \ref{sec:estimation}; finally, we establish a DPO framework with fine-grained preference in section \ref{sec:dpo}.

\subsection{Model Arena Construction}
\label{sec:arena}

In the preference dataset, each entry consists of a pair of images and a binary label. In addition, the model source of each image is recorded. Therefore, the dataset can also be viewed as the result of large-scale pairwise comparisons among multiple models. Based on this observation, we construct an Arena that includes these models, and, as in the K-Sort Arena~\cite{li2025k}, the capability of each model is probabilistically modeled as follows:
\begin{equation}
    \theta \sim \mathcal{N}(\mu, \sigma^2),
\end{equation}
where $\mu$ and $\sigma$ represent the mean and uncertainty of model capability, respectively, and $\mathcal{N}(\cdot)$ denotes the normal distribution.
During each round of pairwise comparison, the current capability distribution $P(\theta)$ of a model is treated as the prior, whereas the binary label defines the likelihood function $P(D | \theta)$ of the observation $D$ given $\theta$.
Based on Bayes’ theorem, the posterior distribution of the capability, denoted as $P(\theta | D)$, is then updated as follows:
\begin{equation}
    P(\theta | D)=\frac{P(D | \theta) P(\theta)}{\int_{-\infty}^{\infty} P\left(D | \theta^{\prime}\right) P\left(\theta^{\prime}\right) d \theta^{\prime}}
    =\frac{P(D | \theta) P(\theta)}{C}.
    \label{eq:post}
\end{equation}
Given the posterior distribution, the mean and variance of the model capability are calculated as follows:
\begin{equation}
\begin{aligned}
\hat{\mu} &= \mathbb{E}[\theta | D]={\textstyle \int_{-\infty}^{\infty} \theta P(\theta | D) d \theta}, \\
\hat{\sigma}^2 &= \operatorname{Var}[\theta | D]={\textstyle \int_{-\infty}^{\infty}(\theta-\mathbb{E}[\theta | D])^2  P(\theta | D) d \theta}.
\label{eq:kwise_update}
\end{aligned}
\end{equation}
The detailed derivations of $\hat{\mu}$ and $\hat{\sigma}^2$ are provided in Appendix \ref{app:Bayesian}.
Based on the above equations, by iterating over the entire dataset, the capability distribution of each model is continuously updated. As proved in K-Sort Arena, $\mu$ and $\sigma$ can eventually converge to values that accurately characterize the true capability distribution of the model.

\subsection{Quality Gap Estimation from Arena}
\label{sec:estimation}
Our goal is to estimate the absolute quality gap between two images in each dataset instance based on the available information, without relying on a reward model. 
Specifically, the available information includes:
(i) the source models’ capability distributions extracted by Arena, parameterized as $\theta^w \sim \mathcal{N}(\mu^w, (\sigma^w)^2)$ and $\theta^l \sim \mathcal{N}(\mu^l, (\sigma^l)^2)$, and (ii) a human-annotated binary preference label $\mathbf{x}_0^{w} \succ \mathbf{x}_0^{l}$.
The quality of each individual image, denoted by $Q^w$ and $Q^l$, is obtained by a single draw from $\theta^w$ and $\theta^l$, respectively. In this case, the problem can be formalized as follows:
\begin{problem}[Arena-based Image Quality Gap Estimation]
Consider two models whose capability distributions are $\theta^w \sim \mathcal{N}(\mu^w, (\sigma^w)^2)$ and $\theta^l \sim \mathcal{N}(\mu^l, (\sigma^l)^2)$. Let $Q^w$ and $Q^l$ be the output qualities obtained by drawing once from $\theta^w$ and $\theta^l$, respectively. Given the observed event $Q^w>Q^l$, calculate the posterior expectation of the quality gap $\mathbb{E}[Q^w-Q^l \mid Q^w>Q^l]$.
\end{problem}
To solve the above problem, we proceed with the following derivation.
Since $Q^w$ and $Q^l$ are independent Gaussian random variables, their difference $Q^D=Q^w-Q^l$ also follows a Gaussian distribution:
\begin{equation}
\begin{gathered}
    Q^D \sim \mathcal{N}(\mu_{Q^D}, \sigma_{Q^D}^2), \\
\text{where} \quad \mu_{Q^D} = \mu^w - \mu^l, \quad 
\sigma_{Q^D}^2 = (\sigma^w)^2 + (\sigma^l)^2.
\end{gathered}
\end{equation}
In each instance of preference datasets, we have the observed outcome $Q^w > Q^l$. This observation is equivalent to conditioning on the event $Q^D > 0$.
Consequently, the estimation of the quality gap reduces to computing the conditional expectation of a Gaussian random variable under a one-sided truncation, i.e., $\mathbb{E}[Q^D \mid Q^D > 0]$.

\begin{theorem}[Expectation the Truncated Normal Distribution]
By the properties of the truncated normal distribution~\cite{johnson1995continuous}, for a random variable $X \sim \mathcal{N}(\mu, \sigma^2)$, the conditional expectation under the constraint $X > a$ admits the following closed-form expression:
\begin{equation}
\begin{gathered}
        \mathbb{E}[X \mid X > a] = \mu + \sigma \, \lambda(\alpha), \\
\text{where} \quad \alpha = \frac{a - \mu}{\sigma}, \quad \lambda(\alpha) = \frac{\phi(\alpha)}{1 - \Phi(\alpha)}.
\end{gathered}
\end{equation}
Here, $\lambda(\alpha)$ is the inverse Mills ratio, and $\phi(\cdot)$ and $\Phi(\cdot)$ denote the probability density function and cumulative distribution function of the standard normal distribution, respectively. The derivation is presented in Appendix \ref{app:Truncated}.
\end{theorem}

\textbf{Latent-Variable Inference via Truncated Normal Distribution }
With Theorem 1, by applying the result to the present case with $X = Q^D$ and $a = 0$, we can obtain the posterior expectation of the quality gap conditioned on the observed comparison as follows:
\begin{equation}
\begin{aligned}
Q^{D*} &=  \mathbb{E}[Q^D \mid Q^D>0]
= \mu_{Q^D} + \sigma_{Q^D} \cdot \frac{\phi\!\left( \frac{-\mu_{Q^D}}{\sigma_{Q^D}} \right)}{1 - \Phi\!\left( \frac{-\mu_{Q^D}}{\sigma_{Q^D}} \right)} \\
&=(\mu^w - \mu^l)
+
\sqrt{(\sigma^w)^2 + (\sigma^l)^2}
\cdot
\frac{
\phi\!\left(
\frac{-(\mu^w - \mu^l)}{\sqrt{(\sigma^w)^2 + (\sigma^l)^2}}
\right)
}{
1 - \Phi\!\left(
\frac{-(\mu^w - \mu^l)}{\sqrt{(\sigma^w)^2 + (\sigma^l)^2}}
\right)
}.
\end{aligned}
\label{eq:gap}
\end{equation}

This result shows that the posterior expected quality gap consists of the prior mean difference augmented by a correction term induced by the selection event $Q^w > Q^l$. 
This quantity exhibits the following properties, making it suitable as a surrogate for the reward model.
When the observed outcome is consistent with the prior ordering (i.e., $\mu^w \gg \mu^l$), the correction term becomes negligible and the posterior expectation is dominated by the prior mean difference. In contrast, when the outcome is unexpected (i.e., $\mu^w < \mu^l$), the correction term becomes dominant, enforcing a strictly positive expected gap consistent with the observed comparison.

\subsection{DPO with Fine-Grained Preference}
\label{sec:dpo}

For each instance, the estimated $Q^{D*}$ is treated as a fine-grained reward signal, enabling DPO to learn more fine-grained preferences. Specifically, instead of only using the ordering between the chosen and rejected responses, we incorporate $Q^{D*}$ as a instance-specific margin that potentially reflects the preference intensity. The resulting BT model with a margin is formulated as follows:
\begin{equation}
    p_{\mathrm{BT}}\left(\mathbf{x}_0^{w} \succ \mathbf{x}_0^{l} \mid \mathbf{c}, Q^D\right)=\sigma\left(r\left(\mathbf{c}, \mathbf{x}_0^{w}\right)-r\left(\mathbf{c}, \mathbf{x}_0^{l}\right)-Q^{D*}\right).
\end{equation}
Then, we integrate this margin-based BT formulation into the Diffusion-DPO~\cite{wallace2024diffusion} training objective, thereby extending standard DPO from enforcing a binary ordering constraint to matching a fine-grained preference signal. The resulting ArenaPO loss is as follows:
\begin{equation}
\begin{aligned}
    \mathcal{L}_{\text {ArenaPO}}(\theta)=-\mathbb{E}_{\mathbf{c},\mathbf{x}_0^w, \mathbf{x}_0^l,t}\Big[  \log \sigma\Big(  
    \beta T  \omega(\lambda_t)
     (\delta_\theta(\mathbf{c}, \mathbf{x}^{w}, t) -\delta_\theta(\mathbf{c}, \mathbf{x}^{l}, t) - \gamma Q^{D*})\Big)\Big],
\end{aligned}
\end{equation}
where $\gamma$ is a coefficient to align the scale of $Q^{D*}$ with $\delta_\theta$.
Compared to standard DPO, which only enforces a positive ordering between image pairs, ArenaPO further encourages the policy to match the magnitude of the preference gap. As a result, the model not only learns which response is better, but also how much better it is, thus achieving more fine-grained and effective preference optimization.

\section{Experiments}

\subsection{Experimental Setup}

\textbf{Models and Datasets }
We perform preference optimization training on Stable Diffusion 1.5~\cite{rombach2022high} and Stable Diffusion XL~\cite{podell2023sdxl}. In addition to the commonly used Pick-a-Pic v2~\cite{pap} dataset, we also perform training on the recent HPD v3~\cite{ma2025hpsv3} dataset, which includes a more diverse set of models and can further leverage the strengths of Arena.

\textbf{Baseline Methods }
We thoroughly compare ArenaPO with existing representative baseline methods for diffusion preference optimization, including Diffusion-DPO~\cite{wallace2024diffusion}, MaPO~\cite{hong2024margin}, DSPO~\cite{zhu2025dspo}, and SDPO~\cite{fu2025diffusion}. Notably, all results reported on the HPD v3 dataset are reproduced using the official open-source implementations with the original hyperparameters.

\textbf{Evaluations }
We adopt multiple preference metrics to verify the performance of trained models, including PickScore~\cite{pap}, HPS v2~\cite{hpsv2}, Aesthetic~\cite{laion-aesthetics}, CLIP~\cite{clip}, and ImageReward~\cite{imagereward}, where the prompts are from Pick-a-Pic v2~\cite{pap}, HPS v2~\cite{hpsv2}, and PartiPrompts~\cite{partiprompts}, respectively. 
We also conduct evaluations on the recent human preference alignment benchmark K-Sort Eval~\cite{ksort-eval}.

\textbf{Implementation Details }
For the Arena construction, we follow the implementation of K-Sort Arena. During training, we incorporate fine-grained preferences into Diffusion-DPO. To ensure a fair comparison, all hyperparameters are kept identical to those used in Diffusion-DPO. Specifically, all models are fine-tuned for 2,000 steps with a global batch size of 2,048, and the learning rates are set to $1\times10^{-8}$ and $1\times10^{-9}$ for Stable Diffusion 1.5 and Stable Diffusion XL, respectively.
After the grid search in section \ref{sec:grid_search}, the hyperparameter $\gamma$ is set to $1\times10^{-3}$ for all experiments.

\subsection{Arena Results}
We first construct a model Arena by treating each image comparison as a competition between the corresponding models. By iterating over the dataset, we obtain the capability distribution $(\mu, \sigma^2)$ of each model. The results on Pick-a-Pic v2 and HPD v3 datasets are in Appendix \ref{app:Arena}.

\begin{wraptable}{R}{0.53\textwidth}\scriptsize
\vspace{-0.4cm}
\centering
\setlength{\tabcolsep}{4pt}
\caption{Estimated quality gaps between two images based on model capabilities in Arena.}
\begin{tabular}{llc}
\toprule
Winning Model ($\mu / \sigma$) & Losing Model ($\mu / \sigma$) & Est. Gap \\ \midrule
\model{FLUX.1-dev}{(28.21/0.80)}         & \model{HunyuanDiT}{(23.41/0.77)}     & 4.8038 \\
\model{Kolors}{(28.54/0.81)}             & \model{Real-Images}{(24.55/0.78)}    & 3.9928 \\
\model{HunyuanDiT}{(23.41/0.77)}         & \model{SDXL}{(21.55/0.78)}           & 1.9662 \\
\model{SD 1-1}{(22.80/0.80)}             & \model{SD 2}{(22.67/0.78)}           & 0.9406 \\
\model{SD 1-4}{(22.97/0.83)}             & \model{VQ-Diffusion}{(26.59/0.82)}   & 0.3211 \\
\bottomrule
\end{tabular}
\label{tab:arena}
\vspace{-0.5cm}
\end{wraptable}

Table \ref{tab:arena} presents examples of estimated quality gaps. When the winning model is intrinsically stronger than the losing model, i.e., its capability mean $\mu$ is higher, the estimated gap is greater than 1 and increases with their difference. Conversely, when the winning model has a lower $\mu$, the gap falls between 0 and 1 and decreases as the difference grows.

\subsection{Main Results of Training on Pick-a-Pic v2 Dataset}
\begin{figure*}[h]
  \centering
  \begin{subfigure}[t]{0.49\textwidth}
    \centering
    \includegraphics[width=\textwidth]{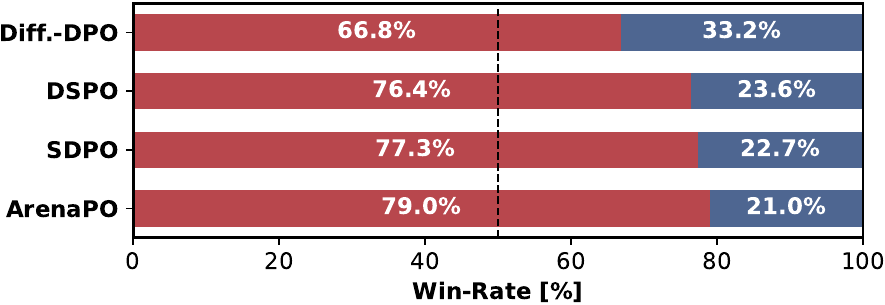}
    \caption{Stable Diffusion 1.5}
  \end{subfigure}
  \hfill
  \begin{subfigure}[t]{0.49\textwidth}
    \centering
    \includegraphics[width=\textwidth]{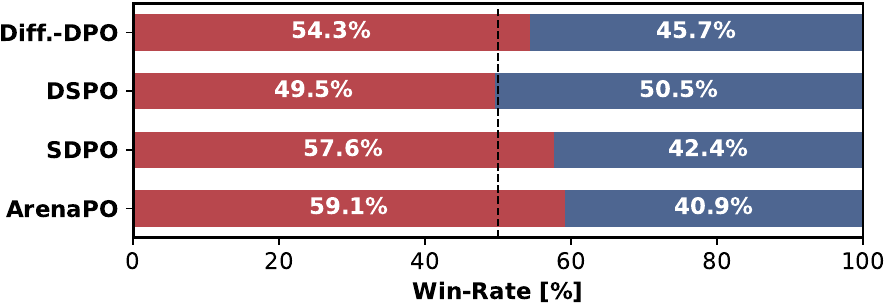}
    \caption{Stable Diffusion XL}
  \end{subfigure}

  \caption{
  Win-rates of models trained using different methods on Pick-a-Pic v2 dataset versus the original model. The test prompts are from HPS v2. The results are the mean values across PickScore, HPS, Aesthetic, CLIP, and ImageReward.
  }
  \label{fig:win-rate}
  \vspace{0cm}
\end{figure*}

The training results on Pick-a-Pic v2 dataset are reported in Table \ref{tab:pap}.
In most cases, ArenaPO achieves the best overall performance across different datasets and backbones. On Pick-a-Pic v2 benchmark with Stable Diffusion 1.5, ArenaPO outperforms all baselines on all five metrics, improving PickScore from 0.2143 (SDPO) to 0.2151 and ImageReward from 0.5546 to 0.6063. 
Moreover, Figure \ref{fig:win-rate} presents the win-rate results, and ArenaPO consistently demonstrates an advantage.

\begin{table*}[h]
\centering
\scriptsize
\setlength{\tabcolsep}{4pt}
\caption{Comparison of preference optimization performance using different methods on Pick-a-Pic v2 dataset. The pretrained backbones include Stable Diffusion 1.5 and Stable Diffusion XL.
}
\begin{tabular}{l|l|ccccc|ccccc}
\toprule
\multirow{2.5}{*}{\textbf{Dataset}} & \multirow{2.5}{*}{\textbf{Method}} & \multicolumn{5}{c|}{\textbf{Stable Diffusion 1.5}} & \multicolumn{5}{c}{\textbf{Stable Diffusion XL}} \\
\cmidrule(lr){3-7} \cmidrule(lr){8-12}
& & \textbf{PickScore} & \textbf{HPS} & \textbf{Aes.} & \textbf{CLIP} & \textbf{IR} & \textbf{PickScore} & \textbf{HPS} & \textbf{Aes.} & \textbf{CLIP} & \textbf{IR} \\
\midrule
\multirow{7}{*}{Pick v2}
&Pre-Trained   & 0.2073 & 0.2651 & 5.3907 & 0.3299 & -0.1376 & 0.2242 & 0.2846 & 5.9970 & 0.3684 & 0.7382  \\
\cmidrule{2-12}
&SFT      & 0.2128 & 0.2765 & 5.6888 & 0.3408 & 0.5767 & 0.2183 & 0.2809 & 5.7922 & 0.3658 & 0.5974 \\
&MaPO & 0.2097 & 0.2702 & 5.5572 & 0.3365 & 0.2435 & 0.2242 & 0.2871 &  \textbf{6.0979} & 0.3684 & 0.8359 \\
&Diff.-DPO      & 0.2109 & 0.2690 & 5.4958 & 0.3357 & 0.1020 & 0.2251 & 0.2868 & 6.0115 & 0.3732 & 0.8357 \\
&DSPO           & 0.2131 & 0.2769 & 5.6825 & 0.3428 & 0.5642 & 0.2228 & 0.2834 & 5.8797 & 0.3756 & 0.8818 \\
&SDPO         & 0.2143 & 0.2772 & 5.7172 & 0.3458 & 0.5546 & 0.2257 & 0.2876 & 5.9812 & 0.3746 & 0.8840 \\
&\cellcolor[HTML]{DFECF6}ArenaPO (ours)  &\cellcolor[HTML]{DFECF6}\textbf{0.2151}    &\cellcolor[HTML]{DFECF6}\textbf{0.2781}    &\cellcolor[HTML]{DFECF6}\textbf{5.7317}   &\cellcolor[HTML]{DFECF6}\textbf{0.3480}  &\cellcolor[HTML]{DFECF6}\textbf{0.6063}    &\cellcolor[HTML]{DFECF6}\textbf{0.2268}      &\cellcolor[HTML]{DFECF6}\textbf{0.2883}      &\cellcolor[HTML]{DFECF6}6.0260      &\cellcolor[HTML]{DFECF6}\textbf{0.3764}      &\cellcolor[HTML]{DFECF6}\textbf{0.9172}      \\
\midrule
\multirow{7}{*}{HPS v2}
&Pre-Trained   & 0.2088 & 0.2697 & 5.4933 & 0.3480 & -0.0469 & 0.2290 & 0.2900 & 6.1271 & 0.3847 & 0.9047 \\
\cmidrule{2-12}
&SFT      & 0.2168 & 0.2838 & 5.7851 & 0.3591 & 0.6619 & 0.2228 & 0.2883 & 5.9689 & 0.3806 & 0.8528 \\
&MaPO & 0.2124 & 0.2760 & 5.6890 & 0.3528 & 0.3308 & 0.2293 & 0.2934 &  \textbf{6.1882} & 0.3840 & 0.9703 \\
&Diff.-DPO      & 0.2131 & 0.2743 & 5.6639 & 0.3552 & 0.1705 & 0.2288 & 0.2927 & 6.1380 & 0.3840 & 1.0159 \\
&DSPO           & 0.2168 & 0.2837 & 5.8346 & 0.3598 & 0.6483 & 0.2273 & 0.2916 & 6.0424 & 0.3894 & 1.0054 \\
&SDPO         & 0.2174 & 0.2827 & \textbf{5.8744} & 0.3600 & 0.6211 & 0.2308 & 0.2938 & 6.1284 & 0.3879 & 1.0326 \\
&\cellcolor[HTML]{DFECF6}ArenaPO (ours)  &\cellcolor[HTML]{DFECF6}\textbf{0.2184}      &\cellcolor[HTML]{DFECF6}\textbf{0.2842}      &\cellcolor[HTML]{DFECF6}5.8591      &\cellcolor[HTML]{DFECF6}\textbf{0.3619}      &\cellcolor[HTML]{DFECF6}\textbf{0.6881}      &\cellcolor[HTML]{DFECF6}\textbf{0.2319}     &\cellcolor[HTML]{DFECF6}\textbf{0.2945}     &\cellcolor[HTML]{DFECF6}6.1685      &\cellcolor[HTML]{DFECF6}\textbf{0.3899}      &\cellcolor[HTML]{DFECF6}\textbf{1.0438}      \\
\midrule
\multirow{7}{*}{PartiPrompts}
&Pre-Trained   & 0.2144 & 0.2724 & 5.3466 & 0.3343 & 0.0637 & 0.2277 & 0.2880 & 5.7901 & 0.3591 & 0.8573 \\
\cmidrule{2-12}
&SFT      & 0.2181 & 0.2821 & 5.5981 & 0.3389 & 0.5830 & 0.2221	& 0.2834 & 5.6496 & 0.3559 & 0.7515 \\
&MaPO & 0.2152 & 0.2754 & 5.4754 & 0.3366 & 0.3358 & 0.2278 & 0.2902 &  5.8921 & 0.3580 & 0.9324 \\
&Diff.-DPO      & 0.2167 & 0.2755 & 5.4045 & 0.3391 & 0.2560 & 0.2279 & 0.2900 & 5.8294 & 0.3629 & 1.0638 \\
&DSPO           & 0.2178 & 0.2819 & 5.5997 & 0.3385 & 0.5640 & 0.2261 & 0.2871 & 5.6947 & 0.3664 & 1.0514 \\
&SDPO         & 0.2187 & 0.2815 & 5.5880 & 0.3423 & 0.5425 & 0.2290 & \textbf{0.2907} & 5.7882 & 0.3645 & 1.0654 \\
&\cellcolor[HTML]{DFECF6}ArenaPO (ours)  &\cellcolor[HTML]{DFECF6}\textbf{0.2198}      &\cellcolor[HTML]{DFECF6}\textbf{0.2833}      &\cellcolor[HTML]{DFECF6}\textbf{5.6164}      &\cellcolor[HTML]{DFECF6}\textbf{0.3445}      &\cellcolor[HTML]{DFECF6}\textbf{0.6291}      &\cellcolor[HTML]{DFECF6}\textbf{0.2299}      &\cellcolor[HTML]{DFECF6}0.2906      &\cellcolor[HTML]{DFECF6}\textbf{5.9003}      &\cellcolor[HTML]{DFECF6}\textbf{0.3671}      &\cellcolor[HTML]{DFECF6}\textbf{1.0672}      \\
\bottomrule
\end{tabular}
\label{tab:pap}
\end{table*}

\subsection{Main Results of Training on HPD v3 Dataset}
To further exploit the strengths of ArenaPO, we also train it on the more recent HPD v3 dataset.
The results are presented in Table \ref{tab:hpd}.
Since this is a newly released dataset, the performance of all baselines is reproduced using the official implementations and hyperparameter settings.
These results demonstrate that ArenaPO consistently achieves the best performance across all three benchmarks. For instance, on HPS v2 with Stable Diffusion XL, ArenaPO pushes PickScore to 0.2328 and ImageReward to 1.0463, which outperforms the previous best SDPO.

\begin{table*}[h]
\centering
\scriptsize
\setlength{\tabcolsep}{4pt}
\caption{Comparison of preference optimization performance using different methods on HPD v3 dataset. The pretrained backbones include Stable Diffusion 1.5 and Stable Diffusion XL. 
}
\begin{tabular}{l|l|ccccc|ccccc}
\toprule
\multirow{2.5}{*}{\textbf{Dataset}} & \multirow{2.5}{*}{\textbf{Method}} & \multicolumn{5}{c|}{\textbf{Stable Diffusion 1.5}} & \multicolumn{5}{c}{\textbf{Stable Diffusion XL}} \\
\cmidrule(lr){3-7} \cmidrule(lr){8-12}
& & \textbf{PickScore} & \textbf{HPS} & \textbf{Aes.} & \textbf{CLIP} & \textbf{IR} & \textbf{PickScore} & \textbf{HPS} & \textbf{Aes.} & \textbf{CLIP} & \textbf{IR} \\
\midrule
\multirow{7}{*}{Pick v2}
&Pre-Trained   & 0.2073 & 0.2651 & 5.3907 & 0.3299 & -0.1376 & 0.2242 & 0.2846 & 5.9970 & 0.3684 & 0.7382  \\
\cmidrule{2-12}
&SFT     & 0.2119  & 0.2773  & 5.6673  & 0.3419   & 0.5852  &  0.2203 &  0.2828 & 5.9062  & 0.3668  &  0.6427 \\
&MaPO & 0.2099  &  0.2760 & 5.6523  &  0.3397 & 0.4894  & 0.2239  & 0.2859  & 5.9862  & 0.3679  &  0.7181 \\
&Diff.-DPO      &  0.2127 & 0.2777  &  5.6628 & 0.3427  & 0.5089  &  0.2255 &  0.2874 & 6.0273  &  0.3739 &  0.8413 \\
&DSPO           &  0.2140 & 0.2782  & 5.7231  & 0.3487  & 0.5825  & 0.2248  &  0.2873 & 6.0288  &  0.3763 & 0.9032  \\
&SDPO        & 0.2145  & 0.2789  &  5.7356 & 0.3482  & 0.5796  &  0.2261 &  0.2879 &  6.0295 & 0.3760  &  0.8983 \\
&\cellcolor[HTML]{DFECF6}ArenaPO (ours)  &\cellcolor[HTML]{DFECF6}\textbf{0.2164}     &\cellcolor[HTML]{DFECF6}\textbf{0.2815}      &\cellcolor[HTML]{DFECF6}\textbf{5.7716}      &\cellcolor[HTML]{DFECF6}\textbf{0.3522}      &\cellcolor[HTML]{DFECF6}\textbf{0.6337}      &\cellcolor[HTML]{DFECF6}\textbf{0.2280}      &\cellcolor[HTML]{DFECF6}\textbf{0.2889}      &\cellcolor[HTML]{DFECF6}\textbf{6.0311}      &\cellcolor[HTML]{DFECF6}\textbf{0.3796}      &\cellcolor[HTML]{DFECF6}\textbf{0.9257}      \\
\midrule
\multirow{7}{*}{HPS v2}
&Pre-Trained  & 0.2088 & 0.2697 & 5.4933 & 0.3480 & -0.0469 & 0.2290 & 0.2900 & 6.1271 & 0.3847 & 0.9047 \\
\cmidrule{2-12}
&SFT        &  0.2189 & 0.2829  &  5.7636 &  0.3573 &  0.6078 & 0.2267  & 0.2879  &  6.1084 & 0.3823  &  0.8749 \\
&MaPO       &  0.2180 & 0.2792  & 5.7521  &  0.3557 &  0.5111 &  0.2286 &  0.2891 & 6.1197  & 0.3840  &  0.8934 \\
&Diff.-DPO      & 0.2188  & 0.2817  & 5.7837  & 0.3613  &  0.5837 & 0.2298  &  0.2941 & 6.1418  &  0.3871 & 1.0192  \\
&DSPO           & 0.2194  &  0.2844 & 5.8728  &  0.3602 & 0.6591  & 0.2309  &  0.2930 &  6.1397 &  0.3867 & 1.0148  \\
&SDPO         &  0.2199 & 0.2851  & 5.8706  &  0.3629 & 0.6628  & 0.2315  & 0.2943  & 6.1572  & 0.3886  & 1.0363  \\
&\cellcolor[HTML]{DFECF6}ArenaPO (ours)  &\cellcolor[HTML]{DFECF6}\textbf{0.2215}      &\cellcolor[HTML]{DFECF6}\textbf{0.2882}      &\cellcolor[HTML]{DFECF6}\textbf{5.8993}      &\cellcolor[HTML]{DFECF6}\textbf{0.3659}      &\cellcolor[HTML]{DFECF6}\textbf{0.6974}      &\cellcolor[HTML]{DFECF6}\textbf{0.2328}      &\cellcolor[HTML]{DFECF6}\textbf{0.2954}      &\cellcolor[HTML]{DFECF6}\textbf{6.1696}      &\cellcolor[HTML]{DFECF6}\textbf{0.3894}      &\cellcolor[HTML]{DFECF6}\textbf{1.0463}      \\
\midrule
\multirow{7}{*}{PartiPrompts}
&Pre-Trained   & 0.2144 & 0.2724 & 5.3466 & 0.3343 & 0.0637 & 0.2277 & 0.2880 & 5.7901 & 0.3591 & 0.8573 \\
\cmidrule{2-12}
&SFT            & 0.2191  &  0.2833 & 5.5862  & 0.3387 &  0.5668 & 0.2263  & 0.2880  & 5.7386  &  0.3538 & 0.8012  \\
&MaPO           &  0.2175 & 0.2798  &  5.5638 & 0.3375  & 0.4729  & 0.2271  & 0.2875  &  5.7876 &  0.3583 &  0.8394 \\
&Diff.-DPO      & 0.2182  & 0.2813  &  5.6210 & 0.3398  &  0.5132 & 0.2290  & 0.2905  & 5.8476  & 0.3628  & 0.9927  \\
&DSPO           &  0.2190 &  0.2831 & 5.6237  & 0.3494  &  0.5825 &  0.2286 & 0.2902  & 5.8521  &  0.3626 & 1.0023  \\
&SDPO         & 0.2194  &  0.2826 & 5.6244  & 0.3513  & 0.5986  & 0.2297  & 0.2910  & 5.8829  & 0.3649  & 1.0274  \\
&\cellcolor[HTML]{DFECF6}ArenaPO (ours)  &\cellcolor[HTML]{DFECF6}\textbf{0.2211}      &\cellcolor[HTML]{DFECF6}\textbf{0.2846}      &\cellcolor[HTML]{DFECF6}\textbf{5.6417}      &\cellcolor[HTML]{DFECF6}\textbf{0.3528}      &\cellcolor[HTML]{DFECF6}\textbf{0.6363}      &\cellcolor[HTML]{DFECF6}\textbf{0.2308}      &\cellcolor[HTML]{DFECF6}\textbf{0.2915}      &\cellcolor[HTML]{DFECF6}\textbf{5.9062}      &\cellcolor[HTML]{DFECF6}\textbf{0.3664}      &\cellcolor[HTML]{DFECF6}\textbf{1.0429}      \\
\bottomrule
\end{tabular}
\label{tab:hpd}
\end{table*}

\subsection{Ablation Studies}

\begin{figure}[t]
  \centering

  \begin{minipage}[t]{0.48\textwidth}
    \centering
    \includegraphics[width=\linewidth]{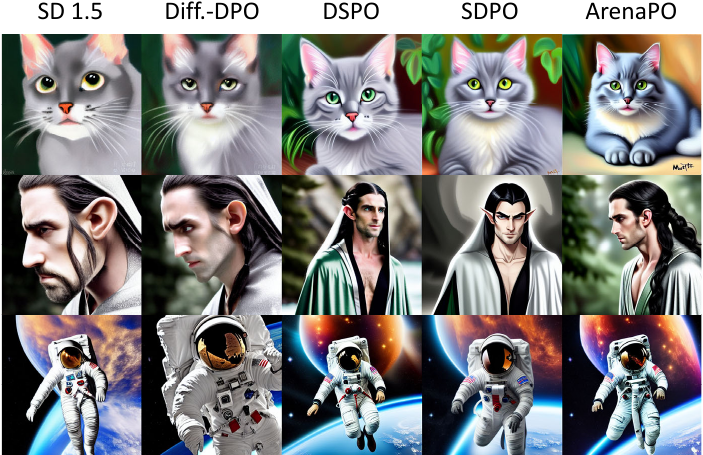}
    \captionof{figure}{Qualitative comparison of different methods for training Stable Diffusion 1.5 on Pick-a-Pic v2 dataset. Prompt: \emph{1) Cute grey cat, digital oil painting by Monet. 2) A pale half-elf in a dark, silver-trimmed robe, with long hair partly tied back. 3) An astronaut floating in space, with Earth dazzling starlight shining in the background.} }
    \label{fig:vis_method}
  \end{minipage}
  \hfill
  \begin{minipage}[t]{0.48\textwidth}
    \centering
    \includegraphics[width=\linewidth]{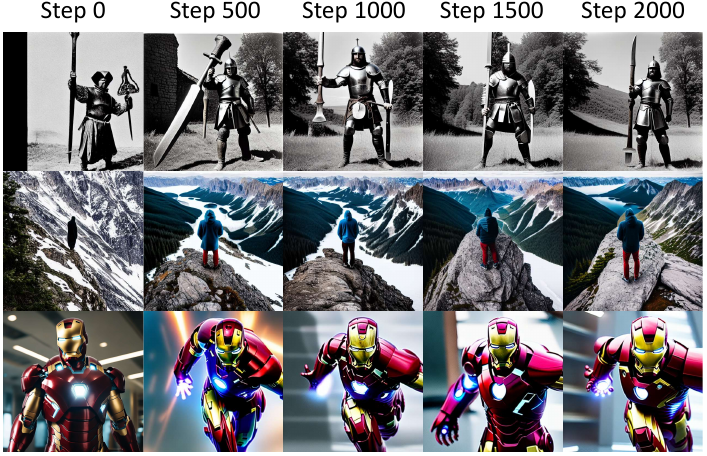}
    \captionof{figure}{Qualitative comparison of different training steps for training Stable Diffusion 1.5 on Pick-a-Pic v2 dataset. Prompt: \emph{1) Photograph of medieval warrior, with giant hammer, posing in battlefield. 2) A person stands on the edge of a cliff, overlooking majestic mountains and a deep valley. 3) The iron man in action, candid shot, capturing the movement.}}
    \label{fig:vis_step}
  \end{minipage}

\end{figure}



\begin{table}[t]
\vspace{-0.3cm}
\centering
\begin{minipage}[t]{0.48\textwidth}
\scriptsize
\centering
\setlength{\tabcolsep}{3.2pt}
\caption{Preference optimization performance of ArenaPO combined with advanced methods. The reported results are obtained by training Stable Diffusion 1.5 on Pick-a-Pic v2 dataset, and the test prompts are from HPS v2.}
\begin{tabular}{l|ccccc}
\toprule
\textbf{Method} & \textbf{PickScore} & \textbf{HPS} & \textbf{Aes.} & \textbf{CLIP} & \textbf{IR} \\
\midrule
Pre-Trained   & 0.2088 & 0.2697 & 5.4933 & 0.3480 & -0.0469 \\
\midrule
SFT      & 0.2168 & 0.2838 & 5.7851 & 0.3591 & 0.6619 \\
MaPO & 0.2124 & 0.2760 & 5.6890 & 0.3528 & 0.3308 \\
Diff.-DPO      & 0.2131 & 0.2743 & 5.6639 & 0.3552 & 0.1705 \\
DSPO           & 0.2168 & 0.2837 & 5.8346 & 0.3598 & 0.6483 \\
SDPO         & 0.2174 & 0.2827 & 5.8744 & 0.3600 & 0.6211 \\
\rowcolor[HTML]{DFECF6}ArenaPO & 0.2184 & 0.2842 & 5.8591 & 0.3619 & 0.6881  \\
\rowcolor[HTML]{C9DFF0}ArenaPO+DSPO & \textbf{0.2191}  &  \textbf{0.2851} &  \textbf{5.8818} &  \textbf{0.3627} &  \textbf{0.6914}    \\
\bottomrule
\end{tabular}
\label{tab:advanced}
\end{minipage}
\hfill
\begin{minipage}[t]{0.48\textwidth}
\scriptsize
\centering
\setlength{\tabcolsep}{4.5pt}
\caption{Comparison of preference optimization performance between ArenaPO and methods with uniform margins. The reported results are obtained by training Stable Diffusion 1.5 on Pick-a-Pic v2 dataset, and the test prompts are from HPS v2.}
\begin{tabular}{l|ccccc}
\toprule
\textbf{Method} & \textbf{PickScore} & \textbf{HPS} & \textbf{Aes.} & \textbf{CLIP} & \textbf{IR} \\
\midrule
Pre-Trained   & 0.2088 & 0.2697 & 5.4933 & 0.3480 & -0.0469 \\
\midrule
Margin=0.5   & 0.2147 & 0.2782 & 5.6976 & 0.3580 & 0.3727 \\
Margin=1.0   & 0.2168 & 0.2830 & 5.8218 & 0.3587 & 0.6287 \\
Margin=2.0   & 0.2174 & 0.2836 & 5.8476 & 0.3602 & 0.6375 \\
Margin=3.0   & 0.2171 & 0.2833 & 5.8525 & 0.3599 & 0.6469 \\
Margin=4.0   & 0.2158 & 0.2811 & 5.7235 & 0.3587 & 0.5724 \\
\rowcolor[HTML]{DFECF6}ArenaPO & \textbf{0.2184} & \textbf{0.2842} & \textbf{5.8591} & \textbf{0.3619} & \textbf{0.6881}    \\
\bottomrule
\end{tabular}
\label{tab:margin}
\end{minipage}
\vspace{-0.3cm}
\end{table}

\textbf{Image Visualization of Different Methods }
Figure \ref{fig:vis_method} presents a qualitative comparison of different methods. Compared to baseline methods, ArenaPO consistently produces images with higher visual quality, better realism, and stronger alignment with human preferences. Please see Appendix \ref{app:vis-1} for more results.

\textbf{Image Visualization of Different Training Steps }
Figure \ref{fig:vis_step} presents a qualitative comparison of different training steps.
We observe that as training proceeds, the generated images exhibit progressively more accurate semantic content and continuously improved visual quality, which verifies the effectiveness of the training process.

\textbf{ArenaPO Combined with Advanced Methods }
Since the offline fine-grained rewards of ArenaPO is decoupled from the DPO optimization strategy, it can be seamlessly integrated with more advanced methods. As shown in Table \ref{tab:advanced}, the combination of ArenaPO and DSPO further yields performance gains, which validates the strong scalability and general applicability of our approach.

\textbf{ArenaPO vs. Uniform Margin }
The reward in ArenaPO is incorporated in the form of a margin. Therefore, we also compare it with fixed-margin baselines, as shown in Table \ref{tab:margin}, where the fixed margin is the replacement of $Q^D$. It can be seen that the uniform margin yields only limited improvements, while ArenaPO provides fine-grained and adaptive awards, resulting in significantly better performance.
To improve the discriminability of the results, we also evaluate on the new benchmark K-Sort Eval~\cite{ksort-eval}, and the results are presented in Appendix \ref{app:ksort-eval}.

\begin{wrapfigure}{r}{0.55\textwidth}
  \vspace{-0.3cm}
  \begin{center}
    \centerline{\includegraphics[width=0.52\columnwidth]{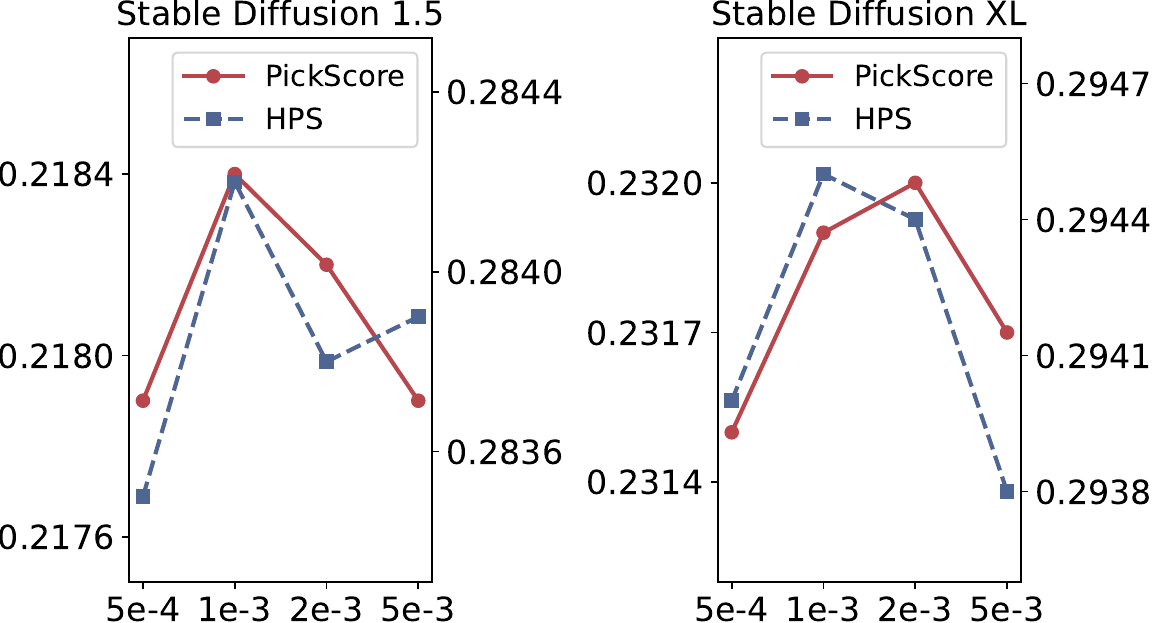}}
    \caption{
      Grid search of the hyperparameter $\gamma$. }
    \label{fig:grid_search}
  \end{center}
    \vspace{-1.5cm}
\end{wrapfigure}

\textbf{Effect of Scaling Coefficient $\gamma$ }
\label{sec:grid_search}
To align the scale when incorporating $Q^D$, we introduce a scaling coefficient $\gamma$. We perform a grid search over $\gamma$, as shown in Figure \ref{fig:grid_search}. The final performance is relatively insensitive to the choice of $\gamma$, and achieves the best overall result at $1\times10^{-3}$. Consequently, for simplicity and general applicability, we fix this parameter to $1\times10^{-3}$ across all experiments.

\section{Conclusion}
In this paper, we propose ArenaPO, an efficient preference optimization framework for diffusion models that extracts fine-grained preference from Arena without relying on an explicit reward model.
By modeling each source model’s capability as a Gaussian distribution, and interpreting binary annotations as observations induced by uncertainty, ArenaPO can estimate quality gaps between image pairs using latent-variable inference via truncated normal distribution.
These fine-grained preferences are then seamlessly integrated into training, combining the efficiency of offline optimization with richer preference signals. 
Extensive experiments on Pick-a-Pic v2 and HPD v3 demonstrate consistent and significant improvements over existing DPO-based methods, validating the effectiveness and practicality of ArenaPO for diffusion model alignment.

\bibliography{ref}
\bibliographystyle{plainnat}

\clearpage
\appendix

\section{Derivation of Bayesian Updating}
\label{app:Bayesian}

Suppose that the observation $D$ indicates that model $M_1$ outperforms model $M_2$. 
The latent performance of model follows a Gaussian distribution, $\theta_i \sim \mathcal{N}(\mu_i, \sigma_i^2)$.
Conditioned on the latent performance variables $\theta_1$ and $\theta_2$, the likelihood of this event is given by:
\[
P(D \mid \theta_1, \theta_2) = P(X_1 > X_2) 
= \Phi\left( \frac{\theta_1 - \theta_2}{\sqrt{\beta_1^2 + \beta_2^2}} \right),
\]
where $\Phi(\cdot)$ and $\phi(\cdot)$ denote the cumulative distribution function (CDF) and the probability density function (PDF) of the standard normal distribution, respectively:
\[
\Phi(x) = \int_{-\infty}^{x} \phi(u)\,du, 
\qquad 
\phi(x) = \frac{1}{\sqrt{2\pi}} e^{-x^2/2}.
\]

Assume independent Gaussian priors $\theta_1 \sim \mathcal{N}(\mu_1, \sigma_1^2)$ and 
$\theta_2 \sim \mathcal{N}(\mu_2, \sigma_2^2)$. 
$\beta$ represents the uncertainty in a model's single-instance performance.
By Bayes' theorem, the joint posterior distribution can be written as:
\[
P(\theta_1, \theta_2 \mid D) \propto 
\phi\!\left(\frac{\theta_1-\mu_1}{\sigma_1}\right)
\phi\!\left(\frac{\theta_2-\mu_2}{\sigma_2}\right)
\Phi\!\left(\frac{\theta_1-\theta_2}{\sqrt{\beta_1^2+\beta_2^2}}\right).
\]

Marginalizing out $\theta_2$ yields the posterior of $\theta_1$:
\[
P(\theta_1 \mid D) \propto 
\phi\!\left( \frac{\theta_1 - \mu_1}{\sigma_1} \right) 
\Phi\!\left( \frac{\theta_1 - \mu_2}{\sqrt{\beta_1^2 + \beta_2^2 + \sigma_2^2}} \right).
\]

The posterior mean of $\theta_1$ is then given by:
\[
\begin{aligned}
\hat{\mu}_1 = \mathbb{E}[\theta_1 \mid D]
&= \mu_1 + \frac{\sigma_1^2}{\sqrt{\sum (\beta_i^2 + \sigma_i^2)}} 
\frac{\phi\!\left(\frac{\mu_1 - \mu_2}{\sqrt{\sum (\beta_i^2 + \sigma_i^2)}}\right)}
{\Phi\!\left(\frac{\mu_1 - \mu_2}{\sqrt{\sum (\beta_i^2 + \sigma_i^2)}}\right)} \\
&= \mu_1 + \frac{\sigma_1^2}{c_{12}} \,
\mathcal{V}\!\left( \frac{\mu_1 - \mu_2}{c_{12}} \right),
\end{aligned}
\]
where
\[
\mathcal{V}(x) = \frac{\phi(x)}{\Phi(x)}, 
\qquad 
c_{ij}^2 = \sum (\beta_i^2 + \sigma_i^2).
\]

Similarly, the posterior variance is updated as:
\[
\begin{aligned}
\hat{\sigma}_1^2 = \mathrm{Var}[\theta_1 \mid D]
&= \sigma_1^2 \left( 1 - \frac{\sigma_1^2}{\sum (\beta_i^2 + \sigma_i^2)} 
\mathcal{W}\!\left( \frac{\mu_1 - \mu_2}{\sqrt{\sum (\beta_i^2 + \sigma_i^2)}} \right) \right) \\
&= \sigma_1^2 \left( 1 - \frac{\sigma_1^2}{c_{12}^2} 
\mathcal{W}\!\left( \frac{\mu_1 - \mu_2}{c_{12}} \right) \right),
\end{aligned}
\]
with
\[
\mathcal{W}(x) = \mathcal{V}(x)\bigl(\mathcal{V}(x) + x\bigr).
\]

This completes the Bayesian update of the performance distribution for a single pairwise comparison in which $M_1$ is observed to outperform $M_2$.

\clearpage

\section{Derivation of Expectation of Truncated Normal Distribution}
\label{app:Truncated}

Let $X \sim \mathcal{N}(\mu, \sigma^2)$ be a Gaussian random variable with probability density function:
\[
f_X(x) = \frac{1}{\sigma} \phi\!\left( \frac{x - \mu}{\sigma} \right),
\]
where $\phi(\cdot)$ denotes the standard normal PDF.

We consider the conditional distribution of $X$ under the truncation constraint $X > a$. 
The conditional density is given by:
\[
f_{X \mid X > a}(x) = 
\frac{f_X(x)}{P(X > a)} 
= \frac{ \frac{1}{\sigma} \phi\!\left( \frac{x - \mu}{\sigma} \right) }
{ 1 - \Phi\!\left( \frac{a - \mu}{\sigma} \right) }, 
\qquad x > a.
\]

The conditional expectation is therefore:
\[
\mathbb{E}[X \mid X > a] 
= \int_a^\infty x \, f_{X \mid X > a}(x)\, dx
= \frac{1}{1 - \Phi(\alpha)} \int_a^\infty x \, \frac{1}{\sigma} \phi\!\left( \frac{x - \mu}{\sigma} \right) dx,
\]
where we define:
\[
\alpha = \frac{a - \mu}{\sigma}.
\]

Apply the change of variables:
\[
z = \frac{x - \mu}{\sigma}, 
\qquad x = \mu + \sigma z, 
\qquad dx = \sigma dz.
\]
Then the integral becomes:
\[
\mathbb{E}[X \mid X > a] 
= \frac{1}{1 - \Phi(\alpha)} 
\int_{\alpha}^{\infty} (\mu + \sigma z) \, \phi(z) \, dz.
\]

Splitting the integral, we obtain:
\[
\mathbb{E}[X \mid X > a]
= \frac{1}{1 - \Phi(\alpha)} 
\left[
\mu \int_{\alpha}^{\infty} \phi(z)\, dz
+ \sigma \int_{\alpha}^{\infty} z \phi(z)\, dz
\right].
\]

Using standard properties of the Gaussian distribution,
\[
\int_{\alpha}^{\infty} \phi(z)\, dz = 1 - \Phi(\alpha),
\]
and
\[
\int_{\alpha}^{\infty} z \phi(z)\, dz = \phi(\alpha),
\]
we arrive at:
\[
\mathbb{E}[X \mid X > a]
= \frac{1}{1 - \Phi(\alpha)} 
\left[
\mu (1 - \Phi(\alpha)) + \sigma \phi(\alpha)
\right].
\]

Finally, simplifying yields:
\[
\mathbb{E}[X \mid X > a] = \mu + \sigma \frac{\phi(\alpha)}{1 - \Phi(\alpha)}.
\]

Defining the inverse Mills ratio:
\[
\lambda(\alpha) = \frac{\phi(\alpha)}{1 - \Phi(\alpha)},
\]
we obtain the desired closed-form expression:
\[
\mathbb{E}[X \mid X > a] = \mu + \sigma \lambda(\alpha).
\]

\clearpage

\section{Arena Results of Pick-a-Pic v2 and HPD v3 Datasets}
\label{app:Arena}

The results of model Arena for Pick-a-Pic v2 and HPD v3 datasets are in Table \ref{tab:arena-app}. It can be observed that Pick-a-Pic v2 primarily emphasizes differences among models with similar architectures, whereas HPD v3 focuses more on comparisons across a diverse set of models.

\begin{table*}[h]
\centering
\small
\caption{Arena results of diffusion models in preference datasets.}
\begin{subtable}{\columnwidth}
\centering
\caption{Pick-a-Pic v2 Training Set}
\setlength{\tabcolsep}{8pt}
\begin{tabular}{lccc}
\toprule
Model & $\mu$ & $\sigma$ & Score \\
\midrule
Stable-Diffusion-XL-beta-v2-2-5-e & 27.99 & 0.78 & 25.66 \\
Stable-Diffusion-XL-beta-v2-2-5-c & 27.32 & 0.77 & 25.00 \\
Stable-Diffusion-XL-beta-v2-2-5-g & 27.26 & 0.77 & 24.95 \\
Stable-Diffusion-XL-beta-v2-2-3-5 & 26.79 & 0.78 & 24.45 \\
Stable-Diffusion-XL-beta-v2-2-5-b & 26.65 & 0.77 & 24.33 \\
Stable-Diffusion-XL-beta-v2-2-5-f & 26.57 & 0.77 & 24.25 \\
Stable-Diffusion-XL-beta-v2-2-5-d & 26.21 & 0.77 & 23.89 \\
Stable-Diffusion-XL-beta-v2-2-5-h & 25.90 & 0.77 & 23.59 \\
Stable-Diffusion-XL-v2-2 & 24.69 & 0.77 & 22.37 \\
Dreamlike-Photoreal-2-flax & 24.54 & 0.77 & 22.24 \\
Stable-Diffusion-XL-beta-v2-2-2 & 24.43 & 0.77 & 22.11 \\
Stable-Diffusion-XL-beta-v2-2-3 & 24.14 & 0.78 & 21.82 \\
Stable-Diffusion-2-1 & 24.08 & 0.78 & 21.75 \\
\bottomrule
\end{tabular}
\end{subtable}

\vspace{0.8em}

\begin{subtable}{\columnwidth}
\centering
\caption{HPD v3 Training Set}
\setlength{\tabcolsep}{12pt}
\begin{tabular}{lccc}
\toprule
Model & $\mu$ & $\sigma$ & Score \\
\midrule
Kolors & 28.54 & 0.81 & 26.11 \\
FLUX.1-dev & 28.21 & 0.80 & 25.82 \\
Infinity & 27.38 & 0.79 & 25.03 \\
VQ-Diffusion & 26.59 & 0.82 & 24.14 \\
Real-Images & 24.55 & 0.78 & 22.21 \\
MidJourney & 23.66 & 0.84 & 21.13 \\
HunyuanDiT & 23.41 & 0.77 & 21.10 \\
Stable-Diffusion-3-Medium & 23.01 & 0.77 & 20.69 \\
Stable-Diffusion-1-4 & 22.97 & 0.83 & 20.48 \\
Stable-Diffusion-1-1 & 22.80 & 0.80 & 20.38 \\
Stable-Diffusion-2 & 22.67 & 0.78 & 20.34 \\
Stable-Diffusion-XL & 21.55 & 0.78 & 19.20 \\
FuseDream & 20.98 & 0.90 & 18.28 \\
CogView2 & 20.11 & 0.80 & 17.71 \\
Glide & 14.91 & 0.87 & 12.31 \\
\bottomrule
\end{tabular}
\end{subtable}
\label{tab:arena-app}
\end{table*}

\clearpage

\section{More Image Visualization of Different Methods }
\label{app:vis-1}

\begin{figure}[h]
  \centering
  \begin{subfigure}[t]{0.48\linewidth}
    \centering
    \includegraphics[width=\linewidth]{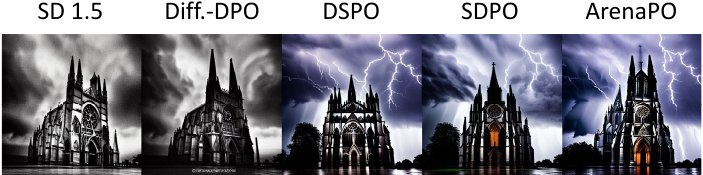}
    \caption{Prompt: \emph{Gothic cathedral in a stormy night.}}
  \end{subfigure}
  \hfill
  \begin{subfigure}[t]{0.48\linewidth}
    \centering
    \includegraphics[width=\linewidth]{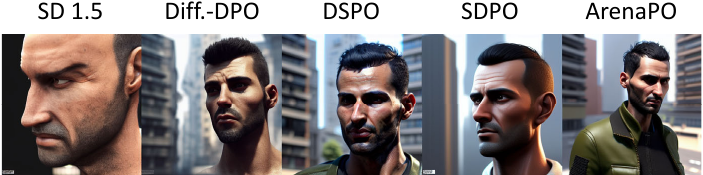}
    \caption{Prompt: \emph{30 year old short slim man, fuller round face, very short hair, black hair, black stubble, olive skin, immense detail/ hyper. }}
  \end{subfigure}

  \begin{subfigure}[t]{0.48\linewidth}
    \centering
    \includegraphics[width=\linewidth]{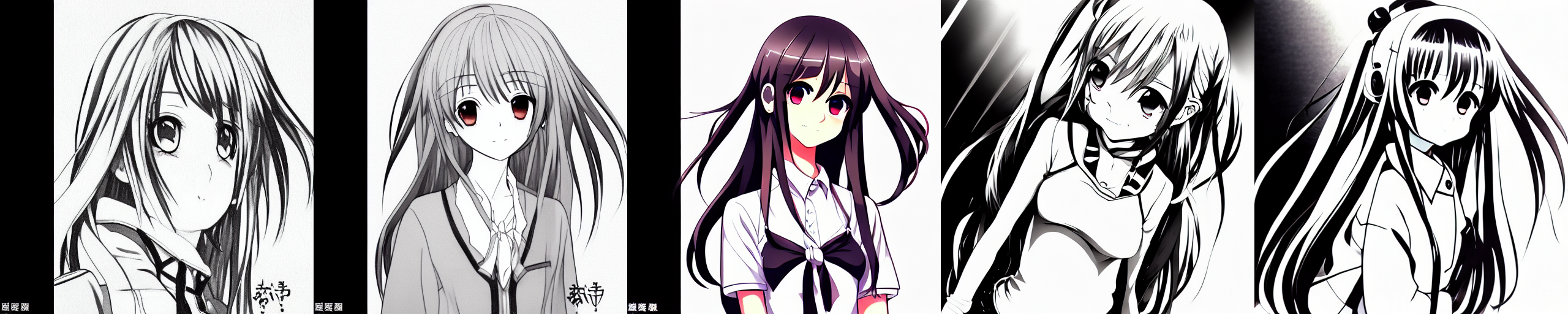}
    \caption{Prompt: \emph{An anime girl, masterpiece, good line art, trending in pixiv.}}
  \end{subfigure}
  \hfill
  \begin{subfigure}[t]{0.48\linewidth}
    \centering
    \includegraphics[width=\linewidth]{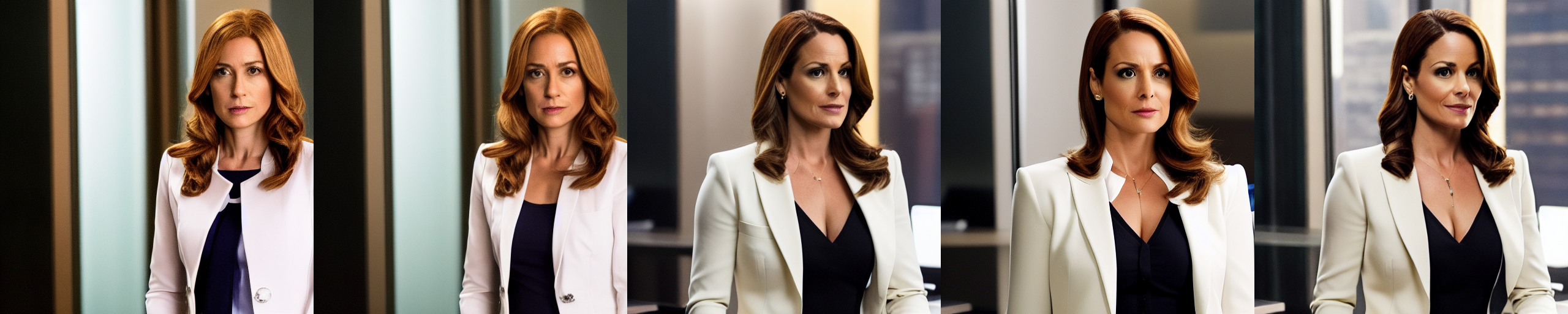}
    \caption{Prompt: \emph{A career woman on suits}}
  \end{subfigure}

  \begin{subfigure}[t]{0.48\linewidth}
    \centering
    \includegraphics[width=\linewidth]{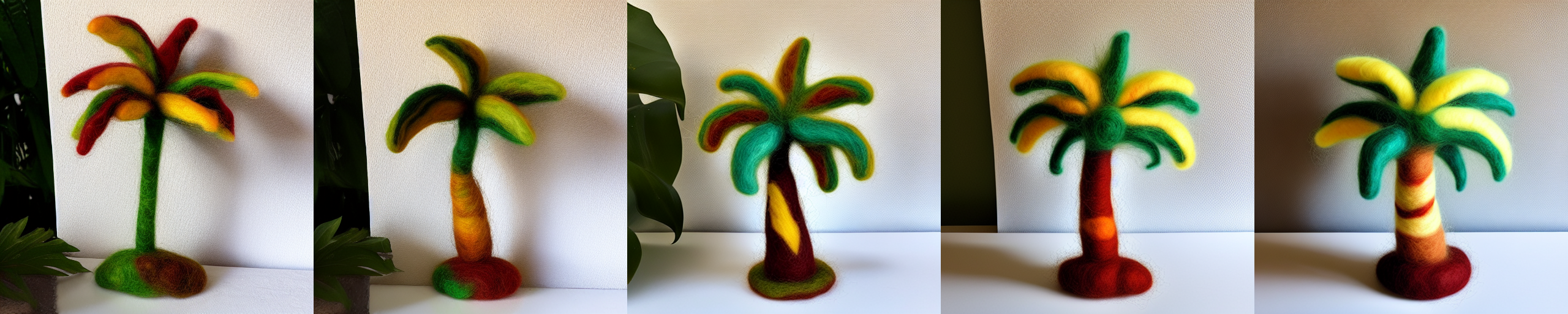}
    \caption{Prompt: \emph{A needle-felted palm tree.}}
  \end{subfigure}
  \hfill
  \begin{subfigure}[t]{0.48\linewidth}
    \centering
    \includegraphics[width=\linewidth]{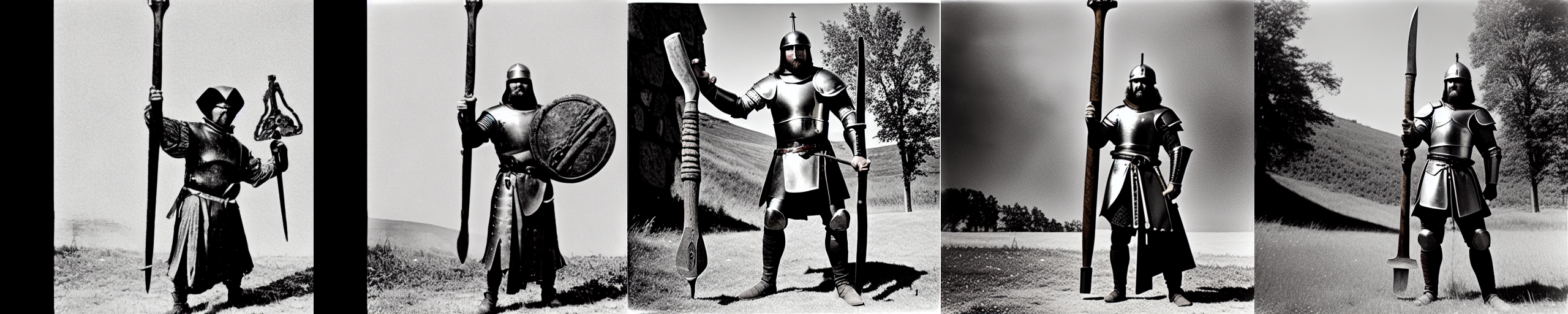}
    \caption{Prompt: \emph{Photograph of medieval warrior, with giant hammer, posing in battlefield.}}
  \end{subfigure}

  \begin{subfigure}[t]{0.48\linewidth}
    \centering
    \includegraphics[width=\linewidth]{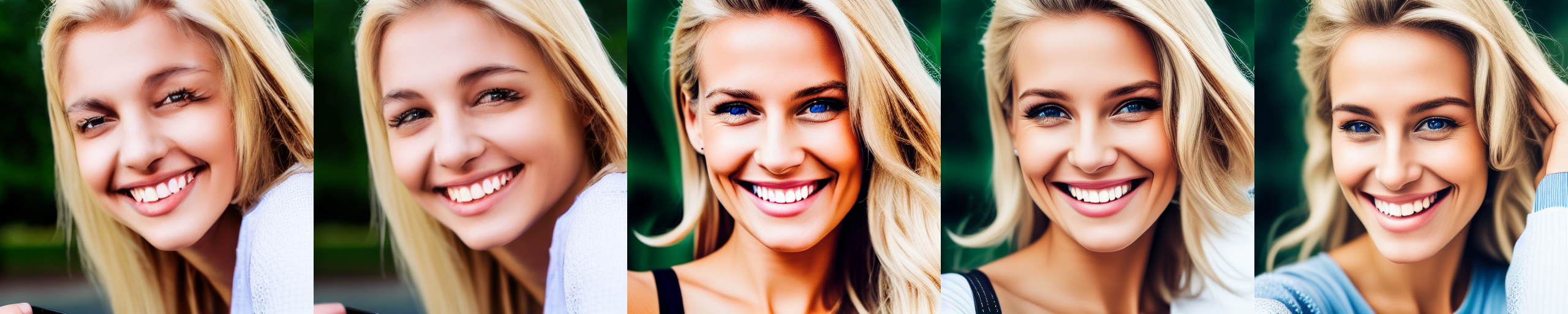}
    \caption{Prompt: \emph{Blonde german-italian girl smiling on a selfie.}}
  \end{subfigure}
  \hfill
  \begin{subfigure}[t]{0.48\linewidth}
    \centering
    \includegraphics[width=\linewidth]{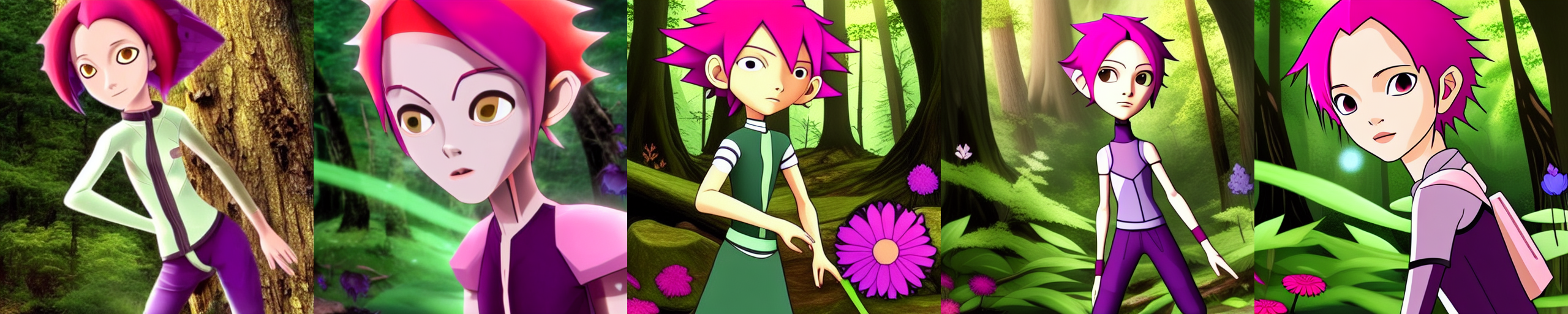}
    \caption{Prompt: \emph{Aelita from Code Lyoko in the forest, gathering flowers.}}
  \end{subfigure}

  \caption{
  More results of qualitative comparison of different methods for training Stable Diffusion 1.5 on Pick-a-Pic v2 dataset.
  }
  \label{fig:vis-app}
\end{figure}

\section{Evaluation on K-Sort Eval Benchmark }
\label{app:ksort-eval}

To ensure a fair comparison, we follow prior work by adopting the same evaluation benchmark and standard metrics. However, as discussed earlier, these traditional metrics are often insufficiently discriminative, making it difficult to clearly distinguish the performance differences among methods.

\begin{wraptable}{r}{0.45\textwidth}\scriptsize
\centering
\caption{Performance comparison on the K-Sort Eval benchmark.}
\label{tab:ksort_eval}
\begin{tabular}{lc}
\toprule
\textbf{Method} & \textbf{K-Sort Eval Score} \\
\midrule
Stable Diffusion 1.5 & 20.03 \\
\midrule
Margin = 0.5 & 20.72 \\
Margin = 2.0 & 21.16 \\
Margin = 4.0 & 20.95 \\
\cellcolor[HTML]{DFECF6}ArenaPO (ours) & \cellcolor[HTML]{DFECF6}\textbf{21.63} \\
\bottomrule
\end{tabular}
\end{wraptable}

To provide a more informative assessment, we additionally evaluate all methods on K-Sort Eval \cite{ksort-eval}, a recently proposed preference-alignment benchmark. K-Sort Eval is an automated evaluation protocol derived from K-Sort Arena and has been shown to correlate well with human judgments.

The evaluation results are summarized in Table~\ref{tab:ksort_eval}. Compared with fixed-margin baselines, ArenaPO consistently achieves the best performance, demonstrating the effectiveness of adaptive preference optimization.

\clearpage

\section{Comparison with RLHF Using an Explicit Reward Model}
A natural question is whether the performance gains of ArenaPO can also be achieved by introducing an explicit reward model, similar to RLHF-style optimization. Following this suggestion, we augment Diffusion-DPO with ImageReward as an explicit reward model and compare it against ArenaPO.

The results are presented in Table~\ref{tab:reward_model}. Incorporating ImageReward substantially improves the IR metric itself, indicating that the model tends to optimize specifically toward the reward model objective. However, improvements on other evaluation metrics remain limited and generally do not surpass ArenaPO. In some cases, such as HPS, the performance even degrades compared with ArenaPO. Similar observations have also been reported in the Diffusion-DPO paper~\cite{wallace2024diffusion}, which notes that \emph{“we did not observe stable improvement when training from public implementations on Pick-a-Pic.”}

\begin{table}[h]\scriptsize
\centering
\caption{Comparison with RLHF-style optimization using ImageReward as an explicit reward model. Best results are shown in bold.}
\label{tab:reward_model}
\begin{tabular}{c|c|ccccc}
\toprule
\multirow{2}{*}{\textbf{Dataset}} & \multirow{2}{*}{\textbf{Method}} 
& \multicolumn{5}{c}{\textbf{Stable Diffusion 1.5}} \\
\cmidrule(lr){3-7}
& & \textbf{PickScore} & \textbf{HPS} & \textbf{Aes.} & \textbf{CLIP} & \textbf{IR} \\
\midrule
\multirow{3}{*}{Pick v2}
& Pre-Trained & 0.2088 & 0.2697 & 5.4933 & 0.3480 & -0.0469 \\
\cmidrule{2-7}
& IR as Reward Model & 0.2162 & 0.2682 & 5.5728 & 0.3556 & \textbf{0.7621} \\
& \cellcolor[HTML]{DFECF6}ArenaPO (ours) & \cellcolor[HTML]{DFECF6}\textbf{0.2184} & \cellcolor[HTML]{DFECF6}\textbf{0.2842} & \cellcolor[HTML]{DFECF6}\textbf{5.8591} & \cellcolor[HTML]{DFECF6}\textbf{0.3619} & \cellcolor[HTML]{DFECF6}0.6881 \\
\bottomrule
\end{tabular}
\end{table}

Moreover, introducing an explicit reward model incurs additional training complexity and computational overhead. Overall, these results suggest that ArenaPO achieves a better trade-off between alignment quality and training efficiency without relying on an external reward model.


\end{document}